\pdfoutput=1

\documentclass[11pt]{article}

\usepackage{ACL2023}
\usepackage{times}
\usepackage{latexsym}
\usepackage{bbm}
\usepackage{bm}

\usepackage[T1]{fontenc}

\usepackage[utf8]{inputenc}
\usepackage{graphicx}
\usepackage{microtype}
\usepackage{booktabs}
\usepackage{multirow}
\usepackage{caption}
\usepackage{subcaption}
\usepackage{amsmath,amssymb}

\usepackage{inconsolata}
\usepackage{color}
\usepackage{paralist}
\usepackage[ruled, vlined, linesnumbered]{algorithm2e}
\usepackage[noend]{algpseudocode}
\usepackage{booktabs}
\usepackage{multirow}
\usepackage{amsmath}
\usepackage{amsthm}
\usepackage{amssymb}
\usepackage{newfloat}
\usepackage{listings}

\usepackage{graphicx}
\usepackage{amsmath}
\usepackage{amssymb}
\usepackage{booktabs}
\usepackage{makecell}
\usepackage{aliascnt}
\usepackage{xcolor}
\usepackage{tcolorbox}
\usepackage{bbm}
\usepackage{tabularx}

\def \tool{SAME\xspace}
\def \model{ $\mathcal{F}$ \xspace}

\title{Dynamic Transformers Provide a False Sense of Efficiency}

\author{Yiming Chen$^{\dag}$ \quad Simin Chen$^{\ddag\thanks{\quad Corresponding author.}}$ \quad Zexin Li$^{\star}$ \quad Wei Yang$^{\ddag}$ \\ \textbf{Cong Liu}$^{\star}$ \quad \textbf{Robby T. Tan}$^{\dag}$ \quad \textbf{Haizhou Li}$^{\natural, \dag, \S}$ \\
        $^\dag$National University of Singapore \quad 
        $^{\natural}$The Chinese University of Hong Kong, Shenzhen \\
        $^{\star}$University of California, Riverside \quad
        $^{\ddag}$University of Texas at Dallas \quad 
        $^{\S}$Kriston AI Lab\\
        \tt yiming.chen@u.nus.edu, \{simin.chen,wei.yang\}@utdallas.edu\\
        \tt \{zli536,congl\}@ucr.edu,
    \{robby.tan,haizhou.li\}@nus.edu.sg \\
}

\begin{document}

\maketitle

\begin{abstract}
Despite much success in natural language processing (NLP), pre-trained language models typically lead to  a high computational cost during inference. Multi-exit is a mainstream approach to address this issue by making a trade-off between efficiency and accuracy, where the saving of computation comes from an early exit. However, whether such saving from early-exiting is robust remains unknown.
Motivated by this, we first show that directly adapting existing adversarial attack approaches targeting model accuracy cannot significantly reduce inference efficiency.
To this end, we propose a simple yet effective attacking framework, \textbf{SAME}, a novel \textbf{s}lowdown \textbf{a}ttack framework on \textbf{m}ulti-\textbf{e}xit models, which is specially tailored to reduce the efficiency of the multi-exit models.
By leveraging the multi-exit models' design characteristics, we utilize all internal predictions 
to guide the adversarial sample generation instead of merely considering the final prediction. 
Experiments on the GLUE benchmark show that SAME can effectively diminish the efficiency gain of various multi-exit models by 80\% on average, convincingly validating its effectiveness and generalization ability.~\footnote{Code is available at \url{github.com/MatthewCYM/SAME}}

\end{abstract}
\section{Introduction}
Pre-trained language models~\citep{devlin-etal-2019-bert,radford2019language,liu2019roberta,lewis-etal-2020-bart,JMLR:v21:20-074, chen2022learning} have shown great potential in a wide range of NLP tasks. While large language models offer unparalleled performance, their high computation during inference limits the scope of applications. 
More studies recently concentrate on efficient NLP, which aims to speed up the inference of deep language models without significant performance degradation~\citep{sanh2019distilbert,zafrir2019q8bert,zhou2020bert}. Among these, the multi-exit models~\citep{zhou2020bert,xin-etal-2020-deebert} attract widespread attention. 

The idea of the multi-exit models stems from the observation that inputs with varying semantics demand distinct computational resources.
By automatically adjusting different computational resources according to input semantics, one can effectively speed up the inference of a multi-exit model with minimum performance loss. Furthermore, such multi-exit model can be easily combined with other static speedup approaches, e.g., distillation~\citep{sanh2019distilbert,jiao-etal-2020-tinybert}, by replacing the backbone model. In addition to higher efficiency, previous studies also show that the multi-exit models are more robust to correctness-based adversarial samples~\citep{zhou2020bert,Hu2020Triple}. 

The study of NLP attacks has mostly focused on harming models' accuracy, and taken static transformers as victim models~\citep{ebrahimi-etal-2018-hotflip,li-etal-2020-bert-attack}. 
There exists another type of attack on the model efficiency, i.e., to make the models computationally slow. 
Considering this type of attack, the intrinsic dynamic nature of the multi-exit models might be vulnerable to such attacks. It remains unexplored, however, how significantly the efficiency or speedup from early exiting will be affected by the attacks.
Motivated by this, we first analyze the efficiency robustness of dynamic NLP transformers. We find that previous accuracy-oriented approaches cannot significantly slow down the dynamic transformers and sometimes even lead to shorter inference time.

To this end, we propose a novel \textbf{s}lowdown \textbf{a}ttack framework on \textbf{m}ulti-\textbf{e}xit language models: \textbf{SAME}. Unlike accuracy-oriented adversarial attacks, there are several unique challenges for effective efficiency attacks. First, existing accuracy-oriented attacks aim to mislead neural networks to generate wrong predictions, which is not suitable for efficiency-oriented attacks. Therefore, we develop a new objective function to guide the generation of efficiency-oriented adversarial samples. 
In addition, our objective function must be general to handle various exit mechanisms in multi-exit transformers.
Second, multi-exit transformers are not static during inference, so the "static" search strategies used in adversarial attacks are not suitable. To overcome the challenges, we propose a dynamic importance adjustment strategy that assigns different importance to each exit layer, allowing the adversarial example search process to focus on the layers that contribute to model efficiency.

We evaluate our SAME using two widely-used multi-exit strategies (entropy-based~\citep{xin-etal-2020-deebert} and patience-based~\citep{zhou2020bert}) with various pre-trained language models~\citep{devlin-etal-2019-bert,liu2019roberta,Lan2020ALBERT} as the backbone on eight tasks from the GLUE benchmark. Experimental results show that our SAME can effectively reduce the computational saving by 80\% on average, which significantly outperforms previous accuracy-oriented approaches by a large margin. Further experiments on the multi-goal attack, attacking transferability, and adversarial training convincingly validate the effectiveness and generalization ability of our proposed SAME.

The contributions of this work are summarised as follows: 
\textit{(1) New Problem:} we identify a new vulnerability of the multi-exit NLP models, namely, the network efficiency.
\textit{(2) Novel Approach:} We propose the first efficiency-oriented attacking framework to measure the efficiency robustness of the multi-exit NLP models.
\textit{(3) Comprehensive Evaluation:} We conduct a systematic evaluation of various dynamic transformers, which shows that future studies on improving and protecting the efficient robustness of the multi-exit NLP models are necessary.

\section{Background}
\subsection{Multi-Exit Networks}
\label{sec:multi-exit-network}

\begin{figure}[ht]
    \centering
    \includegraphics[scale=0.6]{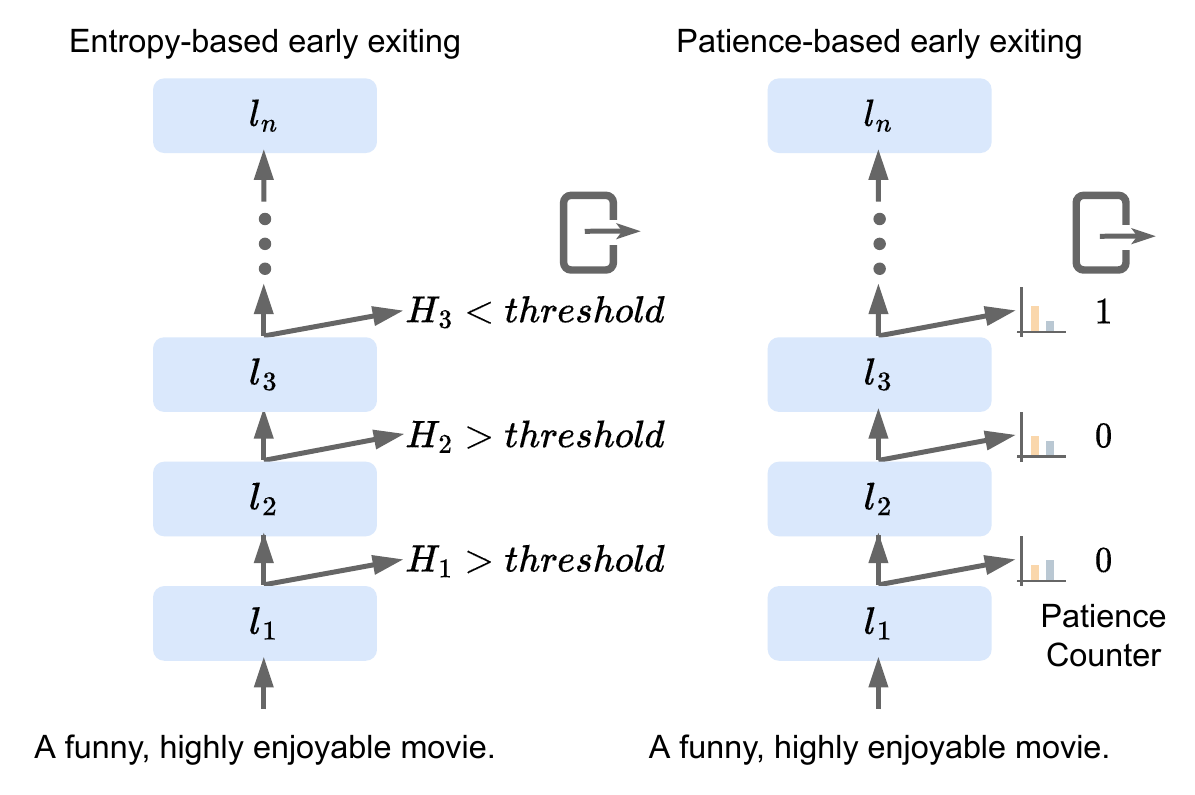}
    \caption{Illustration of entropy-based (left) and patience-based (right) early-exiting strategies, $l_{1...n}$ refer to transformer layers, and $H_{i}$ is the entropy of probability distribution from the $i^{th}$ internal classifier.} 
    \label{fig:ee}
\end{figure}

Multi-exit neural networks include multiple outputs or "exits" placed at different network layers. This architectural design allows for early decision-making if the input is confidently classified or predicted, leading to faster and more efficient processing. 
Based on the semantic complexity of the inputs, multi-exit neural networks can effectively reduce inference time by making predictions from early layers for a simpler input and later layers for a more complex input.
As shown in Figure~\ref{fig:ee}, a multi-exit transformer consists of $N$ transformer layers, each containing an internal classifier.  During the inference phase, predictions are made after each layer, and computation is terminated once the exit criterion is met.  

The choice of exit criterion is crucial in multi-exit models. In this work, we explore two commonly used strategies: entropy-based~\citep{xin-etal-2020-deebert,liu-etal-2020-fastbert} and patience-based~\citep{zhou2020bert,zhu-2021-leebert}. As depicted in Figure~\ref{fig:ee} (left), the entropy-based strategy employs the entropy of a probability distribution as an indicator of model confidence. The model checks if the entropy is lower than a predefined threshold after each layer's computation and outputs a prediction when the criterion is met. The patience-based strategy, as shown in Figure~\ref{fig:ee} (right), involves maintaining a patience counter that is incremented by 1 when predictions from two consecutive internal classifiers are consistent and is reset to zero when they are inconsistent. The model exits early if the patience counter reaches a pre-defined patience threshold.

\subsection{Adversarial Attack}

Adversarial attacks are methods of creating adversarial examples to cause neural networks to make incorrect predictions~\citep{MSH16-FD,ebrahimi-etal-2018-hotflip,LiJDLW19-textbugger,wallace-etal-2019-universal, le-etal-2022-perturbations, hong2020panda,cheng2020seq2sick,li2023sibling, chen2022deepperform, li2023sibling}.
Adversarial attacks in natural language processing (NLP) mainly contain two categories: character-level and word-level.
For the character-level attacks, existing methods involve modifying the words in an input sentence by using insertion, swap, or deletion operators to create adversarial examples~\cite{belinkov2017synthetic,ebrahimi-etal-2018-adversarial}.
The word-level attacks, on the other hand, involve replacing words in the input sentence with other words, e.g., synonym replacement~\citep{ren-etal-2019-generating}, round-trip translation~\citep{zhang-etal-2021-crafting}. There has also been an emergence of attacks targeting generative models. For example, Seq2Sick~\citep{cheng2020seq2sick} generates adversarial examples that decrease the BLUE score of neural machine translation models. In addition to accuracy, inference efficiency is also highly critical for various real-time applications, e.g., speech recognition~\cite{wang2022predict}, machine translation~\citep{fan2021beyond,Zhu2020Incorporating}, lyric transcriptions~\citep{gao2022genre,gao2023polyscriber,gao2022automatic}. Recently, NICGSlowDown and NMTSloth~\citep{chen2022nicgslowdown, chen2022nmtsloth} propose delaying the appearance of the end token to reduce the efficiency of language generative models. 
There have been studies evaluating the accuracy robustness of dynamic transformer through directly adapting TextFooler~\citep{jin2020bert}. Unlike the previous works, the proposed SAME is specially designed for evaluating the efficiency robustness of dynamic transformers.

\section{Methodology}

\subsection{Problem Formulation}

Unlike previous accuracy-oriented approaches, our goal here is to create adversarial examples that decrease the efficiency of a victim multi-exit model \model  by adding human-unnoticeable perturbations to a benign input. Specifically, we focus on two factors: (i) significantly increasing the computational costs for the victim model and (ii) keeping the generated perturbation minimal. We formulate 
the problem as a constrained optimization problem:
\begin{equation}
    \label{eq:problem}
        \Delta = \mathop{argmax}\limits_{\delta} \text{ Exit}_{\mathcal{F}}(x+\delta) 
        \quad  s.t. ||\delta|| \leq \epsilon, 
\end{equation}
where $x$ is the given benign input, $\epsilon$ is the maximum adversarial perturbation allowed, and $\text{Exit}_{\mathcal{F}}(\cdot)$ measures the number of layers where the victim multi-exit language model $\mathcal{F}$ exits. Our proposed approach attempts to find the optimal perturbation $\Delta$ that maximizes the number of layers where the model exits (decrease the efficiency), 
and at the same time adheres to the constraint that the perturbation must be smaller than the allowed threshold (unnoticeable). In this work, we set the allowable modifiable words $\epsilon$ as 10\% of the total input words.

\subsection{Approach Overview}
\begin{figure}
    \centering
    \includegraphics[width=0.48\textwidth]{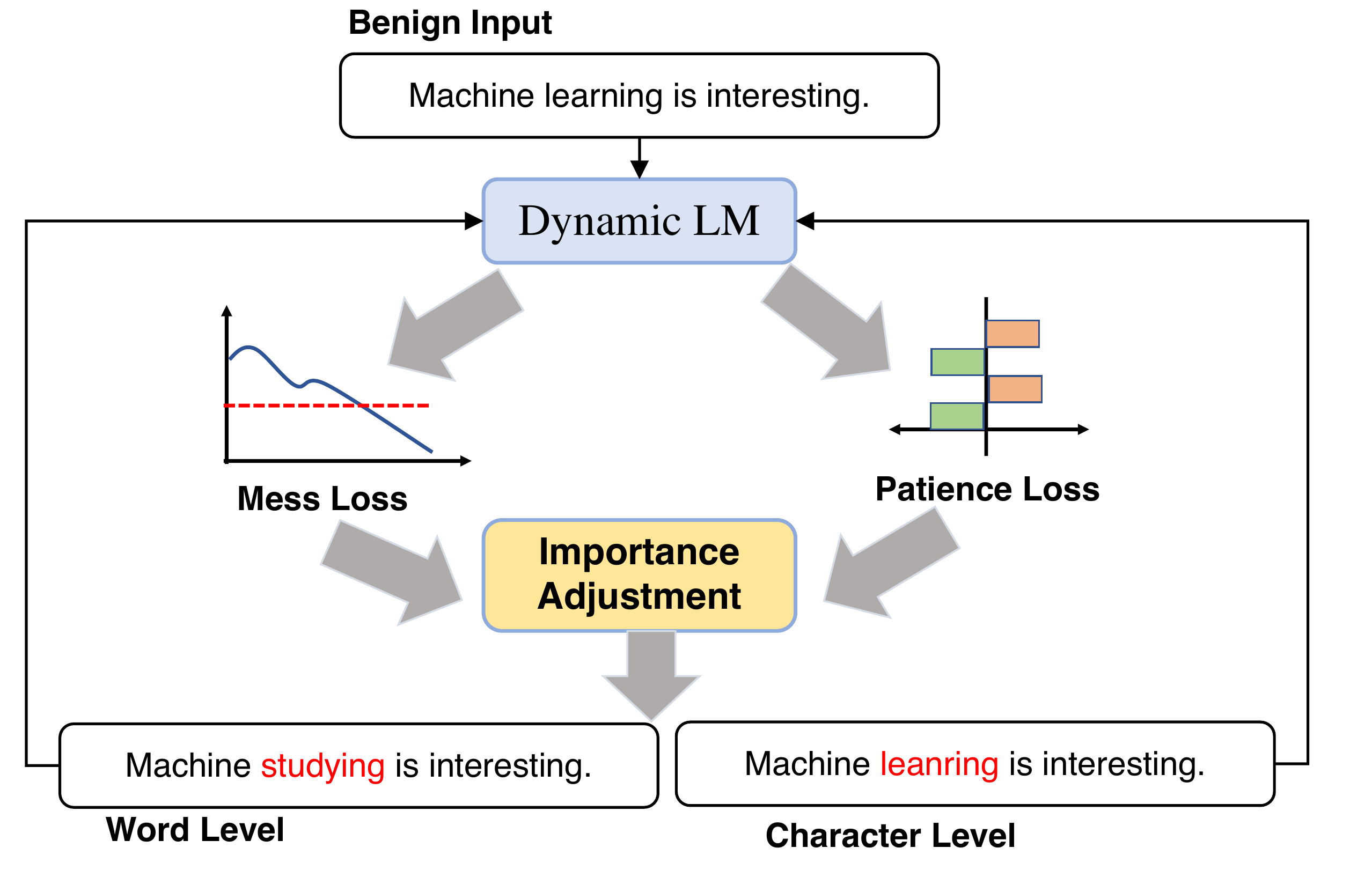}
    \caption{Design overview of SAME }
    \label{fig:overview}
\end{figure}

Figure~\ref{fig:overview} illustrates the design overview of our approach.
Our approach iteratively mutates the given inputs to craft adversarial examples.
During each iteration, we first design a differentiable objective to approximate our adversarial goals (Section~\ref{sec:loss}). 
Then, we dynamically adjust our objective based on the importance of each layer (Section~\ref{sec:dynamic}). 
Finally, we apply our approximated objective function to mutate the inputs with two types of perturbations and generate a set of adversarial candidates that satisfy the given unnoticeable constraints (Section~\ref{sec:perturb}).

\subsection{Adversarial Objective Approximation}
\label{sec:loss}

Notice that our optimization objective in Equation~\ref{eq:problem} is non-differentiable, which makes it challenging to be directly used as the objective for searching optimal adversarial perturbations.
Thus, we need to approximate the adversarial objective (i.e., $argmax \; \text{Exit}_{\mathcal{F}}(\cdot)$) with a differentiable function.
Various objectives are used in accuracy-based adversarial attacks, which aim to decrease the model's accuracy by increasing the confidence scores of the wrong labels. However, these existing approaches do not address the model's efficiency. Therefore, a totally new design for efficiency-oriented adversarial objectives is required. 
Since exiting criteria determine the model's efficiency (as outlined in Section~\ref{sec:multi-exit-network}), we motivate our efficiency-oriented adversarial objective approximation from termination criteria of $\mathcal{F}$, which includes the following:

\textbf{Making Mess Prediction:}
Recall that one way to determine early exiting is by whether the entropy undercuts a predefined threshold. To make the model less efficient, our goal is to keep the entropy above this threshold consistently. It is worth noting that a uniform distribution has the highest entropy among all distributions.
Hence, our first objective function is to push the model prediction close to a uniform distribution:
\begin{equation}
    \label{eq:uniform}
    \mathcal{L}_{mess} = \sum_{i = 1}^N \text{SCE}(\mathcal{F}_i(x), \mathcal{U}),
\end{equation}
where $\mathcal{F}_{i}(x)$ is the prediction logits at the $i^{th}$ layer, $\mathcal{U}$ is a uniform distribution, $N$ is the total layer of the victim $\mathcal{F}$, and SCE$(\cdot)$ is the soft cross entropy loss.
Eq.~\ref{eq:uniform} is interpreted as we seek to minimize the error between output logits (i.e., $\mathcal{F}_i(x)$) and uniform distribution to push the model to produce larger entropy.

\textbf{Decrease Prediction Patience:}
The second termination criterion is based on prediction patience.
To this end, our second objective function needs to push the victim model to produce ``impatient'' predictions.
In other words, we seek to push the model to make inconsistent predictions among its intermediate classifiers as follows:
\begin{equation}
    \label{eq:patience}
    \mathcal{L}_{patience} = \sum_{i = 1} ^N \text{CE}(\mathcal{F}_i(x), h_i),
\end{equation}
where $h_i$ is the constructed target label at the $i^{th}$ layer and CE$(\cdot)$ is the cross entropy function.
As previously mentioned, our second objective seeks to cause the model to produce inconsistent predictions. Thus, we construct our target $h_i$ based as: 
\begin{equation}
    \small
    \label{eq:heur}
             h_i =
        \begin{cases}
        \mathop{argmax}(\mathcal{F}_{i}(x)), & h_{i - 1} \neq  \mathop{argmax}(\mathcal{F}_{i}(x)) \\
        \mathop{argsecond}(\mathcal{F}_{i}(x)), & h_{i - 1} =  \mathop{argmax}(\mathcal{F}_{i}(x))
        \end{cases}, 
\end{equation}
and $h_0$ is set as the prediction given by the model’s first internal classifier on the seed input. Our intuition is to force the model to produce inconsistent predictions between consecutive classifiers by introducing heuristics (Equation~\ref{eq:heur}), thus decreasing prediction patience.

\subsection{Dynamic Importance Adjustment}

\label{sec:dynamic}

It is important to note that the inference path of \model is not ``static'', implying that treating all layer outputs equally at each stage of the search may not yield optimal results. For instance, if \model exits at the third layer, optimizing the input to influence the output before the third layer would be less important. To overcome this challenge, we propose a strategy to dynamically adjust the importance assigned to early layer outputs. Given an input $x$, our layer-wise importance scores are computed as: 
\begin{equation}
    \label{eq:weight}
        w_i =
        \begin{cases}
        \alpha, &i <\text{ Exit}_{\mathcal{F}}(x) \\
        \beta^{i - \text{ Exit}_{\mathcal{F}}(x)}  & i \geq \text{ Exit}_{\mathcal{F}}(x)
        \end{cases}, 
\end{equation}
where $w_i$ is the importance score for the $i^{th}$ layer, $\text{Exit}_{\mathcal{F}}(x)$ is the index of layer that exit the computation, $\alpha$ and $\beta$ are hyper-parameters. 
As shown in Eq.~\ref{eq:weight}, the layers, which have been computed, are assigned constant importance scores, while the layers, which are not used, are assigned exponentially increasing importance scores.

Finally, our objective can be expressed as:
\begin{equation}
\label{eq:total_loss}
            \mathcal{L}_{total} =  \sum_{i=1}\limits^{N}w_i(\lambda \mathcal{L}_{mess}^{i} + (1 - \lambda) \mathcal{L}_{patience}^{i}),
\end{equation}
where $\lambda$ is the hyper-parameters that balance the importance of each objective goals.

\subsection{Perturbing Inputs}
\label{sec:perturb}
Our adversarial perturbation generation includes three main steps: \textit{(i)} finding critical words, \textit{(ii)} generating adversarial candidates, and \textit{(iii)} choosing candidates.

\textbf{Finding Critical Words:} As mentioned earlier, we apply our approximated objective function as guidance to search for optimal adversarial perturbations. Thus, we first find the critical words using the gradient of our objective function (i.e., Equation~\ref{eq:total_loss}).
Specifically, we order the word based on $\mathop{\sum}\limits_j \frac{\partial \mathcal{L}_{total}}{\partial tk_i^j}$, where $tk_i^j$ is the $j^{th}$ dimension of the $i^{th}$ tokens embedding. In this step, we consider the word that is exactly tokenized into one token.

\textbf{Generating Perturbation Candidates:} 
After identifying the critical words, the next step is to perturb the critical words to craft adversarial perturbation candidates.
In this work, we follow existing work and use two types of perturbations to generate adversarial examples: character level and word level, which leads to two variants of SAME: SAME-Char and SAME-Word correspondingly.

For character-level perturbation, we employ four widely used mutations: neighbor character swap, character insertion, character deletion, and homoglyph character replacement~\citep{ebrahimi-etal-2018-adversarial,liu-etal-2022-character}. For neighbor character swap and deletion mutations, we randomly swap or delete one character in the targeted word. To perform character insertion mutation, we randomly select a character from the ASCII character set and then insert it at a random location in the targeted word. For homoglyph character replacement mutation, we use the default homoglyph character mapping from TextBugger~\citep{LiJDLW19-textbugger}. All these four character-level perturbations are common in the real world when typing quickly and can be unnoticeable without careful examination. For each mutation, we randomly generate 25 candidates, resulting in a total of 25$\times$4$=$100 candidates.

For word-level perturbation, we consider replacing the
critical word with another word $\delta$.
To compute the target word, we define word replace increment $I_{s,t}$ to measure the efficiency degradation of replacing word $s$ to $t$:
\begin{equation}
\label{eq:inc}
    \begin{split}
        \mathcal{I}_{s, t} = \sum_{j} (E(t) &  - E(s))_{j} \times \frac{\partial \mathcal{L}_{total}(x)}{ \partial s_i^j};  \\
        \delta = & \mathop{argmax}_{t} \ \mathcal{I}_{s, t}
    \end{split}
\end{equation}
where $E(\cdot)$ represents the embedding vector of a given token, and $\mathcal{I}_{s,t}$ denotes the increase in the direction of the gradient of our objective function, resulting from replacing token $s$ with token $t$.
For word level perturbations, we also generate 100 adversarial candidates.

\textbf{Candidates Selection:} Once the adversarial candidates are generated, we select the valid candidates for the next iteration. To do this, we eliminate candidates that do not meet the constraints in Equation~\ref{eq:problem} and then select the top 5 candidates with the highest $\text{Exit}_\mathcal{F}$ for the next iteration of search.

\begin{table*}[ht]
\centering

\resizebox{\textwidth}{!}{

\begin{tabular}{lcccccccc}
\toprule
\textbf{Method} & \multicolumn{2}{c}{\textbf{SST-2}}& \multicolumn{2}{c}{\textbf{CoLA}}& \multicolumn{2}{c}{\textbf{MRPC}}& \multicolumn{2}{c}{\textbf{QNLI}}\\
\cmidrule(lr){2-3} \cmidrule(lr){4-5} \cmidrule(lr){6-7} \cmidrule(lr){8-9}
 & PD\textless{}2\% & PD\textless{}4\% & PD\textless{}2\% & PD\textless{}4\% & PD\textless{}2\% & PD\textless{}4\% & PD\textless{}2\% & PD\textless{}4\% \\
\midrule
\textbf{DeeBERT-base}        & 2.40x (2.87\%) & 3.06x (0.80\%) & 1.36x (8.72\%)  & 1.40x (5.85\%)  & 1.51x (20.34\%) & 1.98x (6.86\%)  & 1.66x (27.80\%) & 1.83x (20.60\%) \\
\midrule
+HotFlip            & 1.78x (17.09\%) & 2.21x (7.11\%)  & 1.25x (25.98\%)  & 1.27x (22.72\%)  & 1.43x (31.62\%)  & 1.84x (12.75\%)  & 1.54x (32.30\%)  & 1.68x (25.20\%)  \\
+PWWS               & 1.94x (10.32\%) & 2.44x (4.01\%)  & 1.29x (15.63\%)  & 1.31x (12.18\%)  & 1.31x (43.38\%)  & 1.63x (21.32\%)  & 1.46x (39.40\%)  & 1.64x (29.60\%)  \\
+TextBugger         & 1.83x (14.11\%) & 2.25x (5.85\%)  & 1.29x (18.22\%)  & 1.31x (15.72\%)  & 1.27x (51.96\%)  & 1.58x (26.72\%)  & 1.41x (44.10\%)  & 1.52x (36.70\%)  \\
+TextFooler         & 1.83x (15.48\%) & 2.30x (5.62\%)  & 1.29x (15.34\%)  & 1.31x (12.94\%)  & 1.30x (47.55\%)  & 1.65x (18.87\%)  & 1.41x (42.90\%)  & 1.57x (34.40\%)  \\
+A2T                & 2.35x (6.77\%)  & 3.06x (3.10\%)  & 1.29x (22.44\%)  & 1.32x (18.50\%)  & 1.39x (34.31\%)  & 1.80x (12.50\%)  & 1.63x (27.00\%)  & 1.80x (20.70\%)  \\
\midrule
+SAME-Word               & \textbf{1.10x (76.72\%)} & \textbf{1.19x (62.96\%)} & 1.02x (94.15\%)  & 1.02x (92.91\%)  & \textbf{1.02x (94.61\%)}  & 1.09x (77.94\%)  & 1.12x (81.90\%)  & 1.18x (74.40\%)  \\
+SAME-Char & 1.13x (70.87\%) & 1.21x (59.17\%) & \textbf{1.01x (96.84\%)} & \textbf{1.01x (95.69\%)} & 1.02x (93.14\%) & \textbf{1.08x (81.37\%)} & \textbf{1.08x (88.40\%)} & \textbf{1.11x (82.00\%)} \\
\midrule
\textbf{DeeRoBERTa-base}           & 2.31x ( 0.34\%) & 2.74x ( 0.00\%) & 1.29x ( 4.60\%) & 1.30x ( 3.26\%) & 1.63x ( 0.74\%) & 1.86x ( 0.49\%) & 1.48x ( 1.30\%) & 1.61x ( 0.40\%) \\
\midrule
+HotFlip            & 2.28x (1.49\%)  & 2.77x (0.69\%)  & 1.19x (22.15\%) & 1.20x (18.31\%) & 1.58x (2.21\%)  & 1.69x (1.23\%)  & 1.45x (7.90\%)  & 1.61x (2.30\%)  \\
+PWWS               & 2.13x (0.92\%)  & 2.37x (0.34\%)  & 1.21x (14.29\%) & 1.22x (10.93\%) & 1.52x (6.62\%)  & 1.71x (5.39\%)  & 1.40x (7.50\%)  & 1.56x (2.90\%)  \\
+TextBugger & 2.07x (1.95\%)  & 2.29x (0.57\%)  & 1.19x (19.27\%) & 1.20x (15.92\%) & 1.50x (8.33\%)  & 1.61x (8.09\%)  & 1.35x (13.30\%) & 1.50x (4.00\%)  \\
+TextFooler         & 2.09x (1.26\%)  & 2.35x (0.92\%)  & 1.21x (12.08\%) & 1.22x (11.22\%) & 1.52x (6.37\%)  & 1.63x (8.09\%)  & 1.36x (11.40\%) & 1.51x (3.70\%)  \\
+A2T                & 2.34x (0.80\%)  & 2.70x (0.34\%)  & 1.24x (13.23\%) & 1.25x (10.16\%) & 1.59x (3.43\%)  & 1.89x (2.21\%)  & 1.46x (5.70\%)  & 1.63x (1.80\%)  \\
\midrule
+SAME-Word               & 1.63x (16.74\%) & 1.79x (10.09\%) & 1.01x (95.11\%) & 1.01x (94.53\%) & 1.39x (21.57\%) & 1.52x (14.95\%) & 1.23x (31.50\%) & 1.30x (22.30\%) \\
+SAME-Char               & \textbf{1.53x (18.12\%)} & \textbf{1.61x (12.50\%)} & \textbf{1.00x (98.75\%)} & \textbf{1.00x (98.18\%)} & \textbf{1.21x (49.02\%)} & \textbf{1.26x (43.87\%)} & \textbf{1.20x (39.20\%)} & \textbf{1.24x (35.00\%)} \\
\bottomrule
\toprule 
 & \multicolumn{2}{c}{\textbf{QQP}} & \multicolumn{2}{c}{\textbf{RTE}} & \multicolumn{2}{c}{\textbf{MNLI}} & \multicolumn{2}{c}{\textbf{MNLI-mm}} \\
\cmidrule(lr){2-3} \cmidrule(lr){4-5} \cmidrule(lr){6-7} \cmidrule(lr){8-9}
 & PD\textless{}2\% & PD\textless{}4\% & PD\textless{}2\% & PD\textless{}4\% & PD\textless{}2\% & PD\textless{}4\% & PD\textless{}2\% & PD\textless{}4\% \\
\midrule
\textbf{DeeBERT-base}           & 2.68x (3.60\%) & 3.19x (1.40\%) & 1.19x (61.01\%) & 1.71x (14.08\%) & 1.40x (27.80\%) & 1.53x (17.90\%) & 1.42x (25.60\%) & 1.54x (18.00\%) \\
\midrule
+HotFlip            & 2.58x (7.00\%)  & 3.13x (2.60\%)  & 1.21x (59.93\%)  & 1.71x (16.61\%)  & 1.23x (51.40\%)  & 1.34x (37.10\%)  & 1.25x (48.50\%)  & 1.37x (34.20\%)  \\
+PWWS               & 2.77x (3.70\%)  & 3.35x (1.50\%)  & 1.18x (67.15\%)  & 1.54x (31.05\%)  & 1.22x (54.50\%)  & 1.33x (39.10\%)  & 1.24x (53.90\%)  & 1.36x (37.50\%)  \\
+TextBugger      & 2.66x (5.40\%)  & 3.24x (2.40\%)  & 1.18x (66.79\%)  & 1.56x (28.52\%)  & 1.18x (62.70\%)  & 1.30x (43.80\%)  & 1.19x (59.50\%)  & 1.32x (41.80\%)  \\
+TextFooler         & 2.57x (6.30\%)  & 3.19x (2.10\%)  & 1.18x (67.15\%)  & 1.61x (22.74\%)  & 1.20x (56.50\%)  & 1.32x (38.90\%)  & 1.20x (57.00\%)  & 1.34x (38.50\%)  \\
+A2T                & 2.70x (5.20\%)  & 3.35x (1.80\%)  & 1.18x (65.34\%)  & 1.66x (21.66\%)  & 1.30x (39.90\%)  & 1.44x (26.60\%)  & 1.30x (39.30\%)  & 1.43x (25.80\%)  \\
\midrule
+SAME-Word               & \textbf{1.25x (63.10\%)} & \textbf{1.36x (50.60\%)} & \textbf{1.02x (97.11\%)}  & 1.08x (84.48\%)  & 1.01x (96.50\%)  & \textbf{1.01x (96.00\%)}  & \textbf{1.01x (97.70\%)}  & \textbf{1.02x (95.40\%)} \\
+SAME-Char  & 1.37x (55.70\%) & 1.53x (45.50\%) & 1.02x (96.03\%) & \textbf{1.07x (87.00\%)} & \textbf{1.01x (96.90\%)} & 1.02x (94.40\%) & 1.01x (95.70\%) & 1.02x (94.80\%) \\
\midrule
\textbf{DeeRoBERTa-base}           & 2.09x ( 2.30\%) & 2.35x ( 1.40\%) & 1.32x ( 3.97\%) & 1.42x ( 0.00\%) & 1.34x ( 0.70\%) & 1.37x ( 0.40\%) & 1.35x ( 0.40\%) & 1.38x ( 0.30\%) \\
\midrule
HotFlip            & 6.48x (1.30\%)  & 7.78x (0.70\%)  & 1.29x (9.39\%)  & 1.40x (0.00\%)  & 1.31x (5.20\%)  & 1.35x (1.60\%)  & 1.32x (2.80\%)  & 1.37x (1.40\%)  \\
+PWWS               & 2.15x (2.80\%)  & 2.51x (1.30\%)  & 1.28x (12.27\%) & 1.38x (1.08\%)  & 1.32x (3.90\%)  & 1.36x (1.50\%)  & 1.33x (3.20\%)  & 1.37x (1.30\%)  \\
+TextBugger & 2.32x (3.80\%)  & 2.78x (1.60\%)  & 1.26x (17.69\%) & 1.39x (0.00\%)  & 1.31x (6.30\%)  & 1.36x (2.60\%)  & 1.32x (5.20\%)  & 1.37x (2.30\%)  \\
+TextFooler         & 2.08x (5.00\%)  & 2.44x (3.30\%)  & 1.26x (17.69\%) & 1.38x (0.72\%)  & 1.30x (4.70\%)  & 1.35x (1.90\%)  & 1.31x (4.20\%)  & 1.36x (2.00\%)  \\
+A2T                & 2.15x (6.50\%)  & 2.52x (3.40\%)  & 1.29x (10.47\%) & 1.41x (0.00\%)  & 1.33x (3.40\%)  & 1.37x (1.70\%)  & 1.33x (2.90\%)  & 1.38x (1.00\%)  \\
\midrule
+SAME-Word               & \textbf{1.33x (52.90\%)} & \textbf{1.43x (47.50\%)} & 1.14x (49.82\%) & 1.28x (15.88\%) & 1.17x (39.80\%) & 1.19x (34.20\%) & 1.17x (40.00\%) & 1.21x (28.80\%) \\
+SAME-Char               & 1.34x (55.50\%) & 1.47x (47.30\%) & \textbf{1.07x (73.29\%)} & \textbf{1.22x (26.35\%)} & \textbf{1.12x (56.10\%)} & \textbf{1.15x (46.10\%)} & \textbf{1.13x (52.30\%)} & \textbf{1.17x (42.10\%)} \\
\bottomrule
\end{tabular}

}

\caption{Comparison of various attacking methods on entropy-based dynamic models. Attacking methods with lowest speedup are \textbf{bold}.}
\label{tab:main_results_deebert}
\end{table*}

\section{Experiment}
\subsection{Experimental Setup}

\textbf{Datasets:} We conduct our experiments on GLUE~\citep{wang-etal-2018-glue} benchmark. For more details about GLUE, please refer to Appendix \ref{appendix:setup}.

\textbf{Victim models:} We evaluate two popular early-exit strategies, namely entropy-based DeeBERT~\citep{xin-etal-2020-deebert} with backbone model BERT~\citep{devlin-etal-2019-bert} and RoBERTa~\citep{liu2019roberta}, as well as patience-based PABEE~\citep{zhou2020bert} with backbone model BERT~\citep{devlin-etal-2019-bert} and ALBERT~\citep{Lan2020ALBERT}. 
Following the original paper, we consider two different settings with various entropy or patience threshold. Specifically, we select the threshold to keep the relative performance drop within 2\% and 4\%, denoted as PD<2\% and PD<4\%.

\textbf{Baselines:} We compare SAME to 5 recent NLP attack approaches through adapting their attacking strategy to our attacking scenario, which includes white-box attacking approaches: HotFlip~\citep{ebrahimi-etal-2018-hotflip}, TextBugger~\citep{LiJDLW19-textbugger}, A2T~\citep{yoo-qi-2021-towards-improving}; as well as black-box ones: PWWS~\citep{ren-etal-2019-generating}, TextFooler~\citep{jin2020bert}.

\textbf{Metrics:} We evaluate the efficacy of attacking methods with two metrics. As in~\citep{zhou2020bert}, the first metric is the estimated speedup, which is computed as the total number of transformer layers divided by number of actually computed layers. Besides, we propose a second metric, high computation ratio, which refers to the ratio of samples with extremely high computational cost. Specifically, we consider samples with at least 11 computed layers as high computational samples for base-size dynamic transformers with total 12 layers, and at lease 22 computed layers as high computational samples for large-size dynamic transformers with total 24 layers. In all tables, we report the speedup (left) and high computation ratio (right) unless specified otherwise.

\subsection{Main Results}
\label{sec:main_results}
The comparison of different attacking methods on entropy-based dynamic models are shown in Table~\ref{tab:main_results_deebert}, and the results on patience-based models are listed in Table~\ref{tab:main_results_pabee}. Overall, we find that previous accuracy-oriented approaches cannot harm the model efficiency much for either exiting strategy, and even lead to higher speedup for some cases, e.g., QQP, RTE. In sharp contrast, both variants of SAME can effectively reduce the speedup from early exiting, which outperforms all previous approaches by a large margin. Specifically, under PD<2\% setting, SAME eliminates the efficiency gain by 74.88\% on average across GLUE benchmark for DeeBERT series models, and 85\% for PABEE series models. Under PD<4\% setting, model's exiting criteria are more relaxed, which makes the slowdown more difficult. However, SAME consistently reduces the efficiency gain by 75\% for DeeBERT models, and by 82\% for PABEE models, which again convincingly demonstrates the efficacy of SAME.

Besides, while previous works show that patience-based approaches are more robust against accuracy-oriented attack, we observe that both strategies are equally vulnerable under proposed efficiency attack. In addition, these two strategies have different level of vulnerability to different permutation. Entropy-based models are more vulnerable to character-level permutation. On the contrary, word-level permutation performs better on patience-based models. We hypothesize that the discrepancy between two exiting strategies lead to this phenomena. To slowdown patience-based models, ones need to break the consistency between predictions from internal classifiers, which might be difficult to achieve with character-level permutation. The results suggest that further combining multiple level of permutetation methods would lead to a more universal attacking framework that are applicable to wide range of dynamic models.

Finally, we find that the quality of backbone language model has large impact on the efficiency robustness of dynamic transformers. For instance, compared to BERT, RoBERTa is trained with larger corpus with longer time, which makes DeeRoBERTa much more robust than DeeBERT models.

\begin{table*}[ht]
\centering

\resizebox{\textwidth}{!}{

\begin{tabular}{lcccccccc}
\toprule
\textbf{Method} & \multicolumn{2}{c}{\textbf{SST-2}}& \multicolumn{2}{c}{\textbf{CoLA}}& \multicolumn{2}{c}{\textbf{MRPC}}& \multicolumn{2}{c}{\textbf{QNLI}}\\
\cmidrule(lr){2-3} \cmidrule(lr){4-5} \cmidrule(lr){6-7} \cmidrule(lr){8-9}
 & PD\textless{}2\% & PD\textless{}4\% & PD\textless{}2\% & PD\textless{}4\% & PD\textless{}2\% & PD\textless{}4\% & PD\textless{}2\% & PD\textless{}4\% \\
\midrule
\textbf{PABEE-ALBERT-base} & 2.67x (0.00\%) & 3.56x (0.00\%) & 1.63x (9.11\%)  & 1.91x (3.45\%) & 2.03x (1.47\%) & 3.45x (0.00\%) & 1.97x (3.00\%) & 2.41x (1.20\%) \\
\midrule
+HotFlip    & 2.46x (1.38\%)           & 3.30x (0.11\%)           & 1.40x (17.26\%)          & 1.62x (7.86\%)           & 1.78x (7.60\%)           & 3.07x (0.25\%)           & 1.83x (9.20\%)           & 2.27x (3.10\%)           \\
+PWWS       & 2.23x (2.41\%)           & 3.01x (0.23\%)           & 1.40x (11.60\%)          & 1.60x (4.12\%)           & 1.53x (13.48\%)          & 2.57x (0.49\%)           & 1.63x (15.00\%)          & 2.04x (4.50\%)           \\
+TextBugger & 2.20x (2.06\%)           & 2.97x (0.11\%)           & 1.41x (11.12\%)          & 1.60x (3.74\%)           & 1.46x (16.42\%)          & 2.35x (0.49\%)           & 1.48x (22.30\%)          & 1.83x (9.20\%)           \\
+TextFooler & 2.12x (2.64\%)           & 2.96x (0.34\%)           & 1.41x (11.12\%)          & 1.60x (3.93\%)           & 1.47x (19.12\%)          & 2.53x (0.98\%)           & 1.47x (25.80\%)          & 1.85x (10.30\%)          \\
+A2T        & 2.56x (1.03\%)           & 3.51x (0.00\%)           & 1.43x (15.72\%)          & 1.67x (5.18\%)           & 1.80x (9.31\%)           & 3.13x (0.00\%)           & 1.88x (7.70\%)           & 2.33x (2.90\%)           \\
\midrule
+SAME-Word       & \textbf{1.26x (53.10\%) }         & \textbf{1.68x (17.20\%)}          & 1.05x (84.47\%)          & 1.06x (84.95\%)          & \textbf{1.10x (73.28\%)}          & \textbf{1.32x (41.67\%)}          & \textbf{1.28x (50.60\%)}          & \textbf{1.41x (40.90\%)}          \\
+SAME-Char & 1.37x (42.66\%) & 1.77x (14.22\%) & \textbf{1.01x (97.99\%)} & \textbf{1.01x (95.49\%)} & 1.11x (75.25\%) & 1.38x (40.93\%) & 1.30x (49.80\%) & 1.42x (41.80\%) \\
\midrule
\textbf{PABEE-BERT-base}   & 1.66x (9.52\%) & 1.98x (2.41\%) & 1.19x (35.57\%) & 1.19x (35.57\%) & 1.66x (9.56\%)  & 2.01x (2.45\%) & 1.58x (11.00\%) & 1.84x (4.70\%) \\
\midrule
+HotFlip    & 1.49x (22.13\%) & 1.80x (5.50\%)  & 1.05x (80.44\%)  & 1.05x (80.44\%)  & 1.47x (17.40\%)  & 1.79x (1.96\%)  & 1.44x (20.90\%)  & 1.68x (11.40\%) \\
+PWWS       & 1.41x (28.44\%) & 1.66x (10.44\%) & 1.04x (83.70\%)  & 1.04x (83.70\%)  & 1.28x (24.02\%)  & 1.50x (4.41\%)  & 1.33x (32.10\%)  & 1.53x (17.80\%) \\
+TextBugger & 1.37x (32.11\%) & 1.62x (13.53\%) & 1.04x (86.39\%)  & 1.04x (86.39\%)  & 1.25x (30.64\%)  & 1.46x (10.29\%) & 1.25x (44.00\%)  & 1.45x (25.00\%) \\
+TextFooler & 1.37x (32.11\%) & 1.63x (11.12\%) & 1.05x (82.07\%)  & 1.05x (82.07\%)  & 1.26x (28.19\%)  & 1.48x (6.62\%)  & 1.26x (41.70\%)  & 1.45x (21.10\%) \\
+A2T        & 1.62x (12.39\%) & 2.00x (2.98\%)  & 1.07x (72.20\%)  & 1.07x (72.20\%)  & 1.37x (21.32\%)  & 1.66x (3.43\%)  & 1.53x (15.00\%)  & 1.78x (7.60\%)  \\
\midrule
+SAME-Word       &\textbf{ 1.05x (88.19\%)} & \textbf{1.08x (82.11\%)} & \textbf{1.00x (100.00\%)} & \textbf{1.00x (100.00\%)} & \textbf{1.04x (86.76\%)}  &\textbf{ 1.10x (69.85\%)} & \textbf{1.10x (76.00\%)}  & \textbf{1.16x (62.70\%)} \\
+SAME-Char & 1.14x (69.95\%) & 1.21x (61.24\%) & 1.00x (99.90\%) & 1.00x (99.90\%) & 1.05x (85.78\%) & 1.13x (67.16\%) & 1.15x (63.80\%) & 1.23x (54.20\%) \\
\bottomrule
\toprule 
 & \multicolumn{2}{c}{\textbf{QQP}} & \multicolumn{2}{c}{\textbf{RTE}} & \multicolumn{2}{c}{\textbf{MNLI}} & \multicolumn{2}{c}{\textbf{MNLI-mm}} \\
\cmidrule(lr){2-3} \cmidrule(lr){4-5} \cmidrule(lr){6-7} \cmidrule(lr){8-9}
 & PD\textless{}2\% & PD\textless{}4\% & PD\textless{}2\% & PD\textless{}4\% & PD\textless{}2\% & PD\textless{}4\% & PD\textless{}2\% & PD\textless{}4\% \\
\midrule
\textbf{PABEE-ALBERT-base}   & 2.58x (0.50\%)          & 3.40x (0.00\%)          & 1.57x (11.55\%)         & 1.85x (5.05\%)          & 1.86x (4.70\%)          & 2.28x (1.40\%)          & 2.29x (1.60\%)          & 2.29x (1.60\%)          \\
\midrule
+HotFlip    & 2.25x (0.30\%)           & 3.02x (0.00\%)           & 1.46x (20.94\%)          & 1.74x (9.75\%)           & 1.62x (13.30\%)          & 1.99x (4.50\%)           & 2.02x (5.10\%)           & 2.02x (5.10\%)           \\
+PWWS       & 2.34x (1.50\%)           & 3.15x (0.00\%)           & 1.41x (20.58\%)          & 1.61x (11.91\%)          & 1.50x (15.40\%)          & 1.83x (6.20\%)           & 1.81x (5.80\%)           & 1.81x (5.80\%)           \\
+TextBugger & 2.18x (2.30\%)           & 2.88x (0.10\%)           & 1.37x (26.35\%)          & 1.60x (12.27\%)          & 1.45x (20.20\%)          & 1.75x (8.90\%)           & 1.74x (8.20\%)           & 1.74x (8.20\%)           \\
+TextFooler & 2.19x (2.20\%)           & 2.93x (0.50\%)           & 1.37x (28.88\%)          & 1.62x (11.19\%)          & 1.41x (26.00\%)          & 1.72x (11.00\%)          & 1.75x (11.00\%)          & 1.75x (11.00\%)          \\
+A2T        & 2.44x (1.10\%)           & 3.29x (0.20\%)           & 1.45x (22.74\%)          & 1.71x (11.91\%)          & 1.67x (14.90\%)          & 2.06x (5.90\%)           & 2.10x (5.50\%)           & 2.10x (5.50\%)           \\
\midrule
+SAME-Word       & \textbf{1.39x (48.70\%)}          & \textbf{1.65x (27.10\%)}          & 1.13x (67.15\%)          & 1.20x (59.93\%)          & \textbf{1.06x (85.10\%)}          & 1.11x (77.30\%)          & \textbf{1.08x (82.20\%)}          & \textbf{1.08x (82.20\%)}          \\
+SAME-Char & 1.44x (47.90\%) & 1.71x (27.20\%) & \textbf{1.09x (76.90\%)} & \textbf{1.13x (69.68\%)} & 1.06x (86.10\%) & \textbf{1.10x (79.70\%)} & 1.11x (80.40\%) & 1.11x (80.40\%) \\
\midrule
\textbf{PABEE-BERT-base}   & 2.60x (0.40\%) & 3.45x (0.10\%) & 1.21x (55.23\%) & 1.34x (33.21\%) & 1.50x (16.10\%) & 1.76x (7.50\%) & 1.35x (23.30\%) & 1.75x (6.10\%) \\
\midrule
+HotFlip    & 2.45x (1.10\%)  & 3.29x (0.10\%)  & 1.28x (37.91\%)  & 1.42x (28.16\%)  & 1.36x (31.30\%)  & 1.60x (13.90\%) & 1.24x (42.00\%)  & 1.61x (15.30\%) \\
+PWWS       & 2.31x (2.30\%)  & 3.07x (0.00\%)  & 1.26x (44.77\%)  & 1.36x (40.43\%)  & 1.28x (40.40\%)  & 1.49x (18.30\%) & 1.17x (53.20\%)  & 1.50x (16.00\%) \\
+TextBugger & 2.21x (2.80\%)  & 2.78x (1.00\%)  & 1.21x (53.79\%)  & 1.34x (39.35\%)  & 1.27x (42.60\%)  & 1.45x (22.00\%) & 1.15x (57.00\%)  & 1.44x (23.50\%) \\
+TextFooler & 2.16x (3.70\%)  & 2.91x (0.30\%)  & 1.27x (44.40\%)  & 1.43x (31.05\%)  & 1.22x (52.50\%)  & 1.44x (23.10\%) & 1.13x (63.50\%)  & 1.42x (26.20\%) \\
+A2T        & 2.54x (0.90\%)  & 3.47x (0.10\%)  & 1.29x (40.43\%)  & 1.45x (27.80\%)  & 1.41x (29.00\%)  & 1.66x (13.20\%) & 1.25x (41.60\%)  & 1.63x (15.00\%) \\
\midrule
+SAME-Word       & \textbf{1.35x (49.60\%)} & \textbf{1.62x (24.90\%)} & 1.09x (75.45\%)  & 1.14x (71.12\%)  & \textbf{1.01x (96.20\%)}  & \textbf{1.03x (93.00\%)} & \textbf{1.01x (97.80\%)}  & \textbf{1.03x (92.70\%)} \\
+SAME-Char & 1.58x (32.90\%) & 1.95x (11.20\%) & \textbf{1.08x (74.01\%)} & \textbf{1.11x (71.84\%)} & 1.03x (91.80\%) & 1.06x (85.90\%) & 1.02x (93.50\%) & 1.06x (86.70\%) \\
\bottomrule
\end{tabular}

}

\caption{Comparison of various attacking methods on patience-based dynamic models. Since patience threshold is a discrete number, some entries share the same value, e.g., PD<2\% and PD<4\% for PABEE-BERT on CoLA.}
\label{tab:main_results_pabee}
\end{table*}

\subsection{Accuracy \& Efficiency}
Since another important adversarial goal is misclassification, we further investigate the trade-off between accuracy and efficiency drop during attacking. Table~\ref{tab:acc_results} summarizes the results on SST-2 and MNLI-mm. In addition to efficiency drop, SAME can also considerably lead to misclassification. As the goal function of SAME doesn't consider the accuracy metric, we further propose SAME+, which adopts a multi-objective goal function:
\begin{equation}
    \text{ Exit}_{\mathcal{F}}(x+\delta) + \sigma \times \mathbbm{1}(\mathcal{F}(x+\delta)\neq y_{true}),
\end{equation}
where $y_{true}$ is the ground truth label, $\mathbbm{1}(\cdot)$ is the indicator function, and $\sigma$ is the weight that balances the importance of accuracy and efficiency. As we focus on efficiency robustness in this work, we set $\sigma$ to 0.5. Therefore, SAME+ is expected to produce adversarial samples with a similar efficiency drop level as SAME but leads to an additional accuracy drop. As shown in Table~\ref{tab:acc_results}, the average accuracy score can be further reduced by 42.26\% for SAME-word and 37.47\% for SAME-char without any increase in efficiency. In addition, previous work shows that patience-based methods are more robust against accuracy-oriented adversarial attack, compared to entropy/confidence-based ones~\citep{zhou2020bert}. However, we observe that SAME leads to similar accuracy drop for patience-based and entropy-based dynamic models. The robustness of patience-based methods come from internal classifier ensemble. Yet, proposed heuristic loss in SAME makes these internal classifiers hard to reach an agreement. Then, the victim model will directly obtain prediction from the last classifier for large proportion of inputs, which actually fails the mechanism of internal classifier ensemble.
The empirical results suggest that it's possible to craft adversarial samples with low accuracy and efficiency.

\begin{table}[ht]
\centering
\resizebox{\columnwidth}{!}{

\begin{tabular}{lcccc}
\toprule
\textbf{Method} & \multicolumn{2}{c}{\textbf{SST-2}}  & \multicolumn{2}{c}{\textbf{MNLI-mm}}    \\
\cmidrule(lr){2-3} \cmidrule(lr){4-5}
 & DeeBERT & PABEE-BERT & DeeBERT & PABEE-BERT \\
\midrule
DeeBERT-base    & 89.91 (2.40x) & 90.83 (1.66x) & 85.40 (1.42x) & 82.40 (1.35x) \\
\midrule
+SAME-Word       & 63.53 (1.10x) & 64.22 (1.05x) & 53.00 (1.01x) & 58.40 (1.01x) \\
+SAME-Char       & 71.22 (1.13x) & 73.85 (1.14x) & 59.50 (1.01x) & 61.20 (1.02x) \\
\midrule
+SAME-Word+      & 24.08 (1.10x) & 20.07 (1.05x) & 7.60 (1.01x) & 4.60 (1.00x) \\
+SAME-Char+      & 40.37 (1.14x) & 45.41 (1.14x) & 13.90 (1.02x) & 16.20 (1.02x) \\
\bottomrule
\end{tabular}
}

\caption{Comparison of SAME with(out) accuracy multi-goal function: each entry gives accuracy (left) and speedup (right).}
\label{tab:acc_results}
\end{table}

\subsection{Attacking Transferability}
In this section, we examine whether adversarial samples from SAME are transferable between various architectures. We study two settings: \textbf{(i) Cross backbone:} we assume the source model and target model share the same early exiting strategy but with different backbone models. \textbf{(ii) Cross mechanism:} we assume that the source and target model have different early exiting strategies.

Table~\ref{tab:transfer} summarizes the results on SST-2 and MNLI datasets. Overall, the adversarial samples are transferable between different models, and several critical factors determine the transferability. The first one is the exiting strategy. We find that samples are more transferable between models sharing the same exiting strategy, e.g., from PABEE-ALBERT-base to PABEE-BERT-base. The second factor is the backbone model. If the source and target model have the same backbone language model or share the same tokenizer, e.g., DeeBERT-base and DeeBERT-large, the transferred samples will cause more slowdown. In addition, we find that entropy-based models are more vulnerable to transferred attacks compared to patience-based models. Interestingly, we again observe that character-level attack is more transferable to the entropy-based model. while the word-level attack is more transferable to the patience-based model, which is consistent with our findings from Section~\ref{sec:main_results}.

\begin{table}[ht]
\centering

\resizebox{0.46\textwidth}{!}{

\begin{tabular}{lcccc}
\toprule
\textbf{Source model} & \multicolumn{2}{c}{\textbf{SST-2}} & \multicolumn{2}{c}{\textbf{MNLI}} \\
\cmidrule(lr){2-3} \cmidrule(lr){4-5}
            & Char        & Word       & Char   & Word  \\
\midrule
 & \multicolumn{4}{c}{DeeBERT-base} \\
\midrule
DeeBERT-large                & 49.51\%          & 33.98\%         & 52.83\%     & 41.51\%    \\
DeeRoBERTa-base              & 35.92\%          & 17.96\%         & 37.74\%     & 29.59\%    \\
PABEE-ALBERT-base             & 37.86\% & 10.19\% & 45.28\% & 37.74\%  \\
PABEE-BERT-base              & 50.49\% & 27.67\% & 54.72\% & 45.28\%   \\
\midrule
& \multicolumn{4}{c}{PABEE-BERT-base} \\
\midrule
PABEE-ALBERT-base & 17.35\% & 23.47\% & 28.95\% & 30.26\%   \\
PABEE-BERT-large   & 28.57\% & 35.71\% & 32.89\% & 32.89\%   \\
DeeBERT-base      & 31.63\% & 38.78\% & 27.63\% & 34.21\%   \\
DeeRoBERTa-base   & 12.24\% & 19.39\% & 15.78\% & 17.11\%  \\
\bottomrule
\end{tabular}
}
\caption{Transferability results: the first block shows the results with DeeBERT as the target model, and the second block uses PABEE-BERT as the target model. Each row refers to a different source model. Char and Word refer to varients of SAME. Each entry denotes the efficiency gain decrease ratio.}
\label{tab:transfer}
\end{table}

\subsection{Adversarial Training}
We further explore whether this new efficiency threat can be successfully defended through adversarial training. Specifically, given a victim model. we first generate an adversarial sample using SAME or other adversarial approaches for each sample from the training set. Then, we equally mix the clean and adversarial samples to retrain a new model. Finally, we attack the adversarial trained models again with SAME. We adjust the entropy/patience of adversarial trained models to have the same speedup as the original victim model. Table~\ref{tab:adv_train_results} shows the results. Overall, the efficiency robustness of dynamic transformers can be improved through adversarial training (1.18x to 1.58x on average using TextFooler), Yet, there still exists a drastic speedup loss (2.25x to 1.58x). Compared to accuracy-oriented adversarial data, data from SAME provide more robustness beneficial against attack, which validates the potential of using SAME to enhance the robustness of current dynamic transformers. 

\begin{table}[ht]
\centering

\resizebox{\columnwidth}{!}{

\begin{tabular}{lcccc}
\toprule
\textbf{Method} & \multicolumn{2}{c}{\textbf{MRPC}}  & \multicolumn{2}{c}{\textbf{RTE}}    \\
\cmidrule(lr){2-3} \cmidrule(lr){4-5}
 & DeeBERT & PABEE-ALBERT & DeeBERT & PABEE-ALBERT \\
\midrule

Clean & 1.98x ( 6.86\%) & 3.45x ( 0.00\%) & 1.71x ( 14.08\%) & 1.85x ( 5.05\%) \\
\midrule
w/o AdvTrain  & 1.09x (77.94\%) & 1.32x (41.67\%)  & 1.08x (84.48\%) & 1.20x (59.93\%) \\
TextFooler        & 1.39x (45.83\%) & 2.30x (5.88\%) & 1.26x (37.55\%) & 1.37x (34.66\%) \\
PWWS              & 1.32x (38.73\%) & 1.87x (8.09\%) & 1.09x (77.98\%) & 1.41x (23.10\%) \\
SAME              & 1.36x (34.80\%) & 3.20x (0.74\%) & 1.22x (57.40\%) & 1.34x (39.71\%) \\
\bottomrule
\end{tabular}
}

\caption{Efficiency of models trained with various adversarial augmented data. Each row refers to a model trained with different adversarial data.}
\label{tab:adv_train_results}
\end{table}

\subsection{Discussion}
\textbf{Impact of Model Scale:} Since attacking approaches is required to slowdown the victim models by more layers to achieve the same slowdown ratio, we further investigate the impact of victim model scale on the attacking performance. Experimental results using 24-layer BERT-large model on SST-2 and MNLI are shown in Table~\ref{tab:large_results}. Due to space limitation, more results can be found in Appendix~\ref{appendix:large}. Accuracy-oriented methods can still hardly reduce the inference efficiency. Yet, our proposed SAME effectively reduce the speedup ratio by 89\%, which is comparable to 93\% on base-size models.

\begin{table}[ht]
\centering

\resizebox{\columnwidth}{!}{

\begin{tabular}{lcccc}
\toprule
\textbf{Method} & \multicolumn{2}{c}{\textbf{SST-2}}  & \multicolumn{2}{c}{\textbf{MNLI}}    \\
\cmidrule(lr){2-3} \cmidrule(lr){4-5}
 & DeeBERT & PABEE-BERT & DeeBERT & PABEE-BERT \\
\midrule
w/o AdvTrain      & 2.06x (2.06\%) & 2.91x (0.57\%) & 1.57x (1.50\%) & 1.56x (15.20\%) \\
\midrule
+HotFlip            & 1.74x (12.73\%) & 2.53x (1.72\%) & 1.45x (7.10\%) & 1.40x (29.70\%)  \\
+PWWS               & 1.91x (7.11\%)  & 2.38x (2.41\%) & 1.45x (4.00\%) & 1.26x (44.80\%) \\
+TextBugger         & 1.87x (6.88\%)  & 2.35x (1.95\%) & 1.43x (6.90\%) & 1.23x (50.00\%)  \\
+TextFooler         & 1.92x (8.37\%)  & 2.32x (1.95\%) & 1.43x (8.40\%) & 1.23x (49.40\%)  \\
+A2T                & 2.18x (4.93\%)  & 2.79x (1.03\%) & 1.51x (6.20\%) & 1.42x (29.50\%)  \\
\midrule
+SAME               & \textbf{1.11x (65.71\%)} & \textbf{1.22x (58.49\%)} & \textbf{1.05x (81.90\%)} & \textbf{1.04x (89.50\%)} \\
\bottomrule
\end{tabular}
}

\caption{Attacking results on large dynamic transformers with 24 transformer layers.}
\label{tab:large_results}
\end{table}

\textbf{Impact of modification rate:} In our main results, we set the allowable modification rate $\epsilon$ as 10\% of the input words. We further investigate whether SAME can reduce the inference efficiency under lower modification rate (imperceptible attack). The experiment results across GLUE benchmark on DeeBERT-base and PABEE-BERT-base under are summarized in Table~\ref{tab:modification_rate_results}. Even constrained with a very low modification rate, e.g., 3\%, both variants of SAME can still significantly reduce the model's efficiency. In addition, with increasing modification rate, SAME leads to higher reduction in efficiency.

\begin{table}[ht]
\centering

\resizebox{\columnwidth}{!}{

\begin{tabular}{lcccccccc}
\toprule
$\bm{\epsilon}$ & \multicolumn{4}{c}{\textbf{DeeBERT}}                                      & \multicolumn{4}{c}{\textbf{PABEE-BERT}}                                   \\
\cmidrule(lr){2-5} \cmidrule(lr){6-9}
           & \multicolumn{2}{c}{Word}            & \multicolumn{2}{c}{Char}            & \multicolumn{2}{c}{Word}            & \multicolumn{2}{c}{Char}            \\
\cmidrule(lr){2-3} \cmidrule(lr){4-5} \cmidrule(lr){6-7} \cmidrule(lr){8-9}
                    & PD\textless{}2\% & PD\textless{}4\% & PD\textless{}2\% & PD\textless{}4\% & PD\textless{}2\% & PD\textless{}4\% & PD\textless{}2\% & PD\textless{}4\% \\
\midrule
3\%                 & 70.46            & 67.84            & 70.64            & 68.33            & 69.68            & 66.26            & 60.63            & 58.61            \\
5\%                 & 83.81            & 80.70            & 82.56            & 81.67            & 81.05            & 76.78            & 71.79            & 70.22            \\
7\%                 & 85.41            & 83.50            & 84.52            & 83.37            & 82.53            & 79.78            & 73.68            & 71.04            \\
10\%                & 90.21            & 88.47            & 88.43            & 87.26            & 86.32            & 84.15            & 77.89            & 76.09           \\
\bottomrule
\end{tabular}
}

\caption{Average efficiency reduction ratio on GLUE benchmark under various modification rate $\epsilon$.}
\label{tab:modification_rate_results}
\end{table}

\textbf{Ablation Study:} To understand the inner mechanism of \tool, we conduct ablation studies on each component. As shown in Table~\ref{tab:ablation}, solely using heuristic loss can already lead to significant efficiency drop. In addition, using loss combination, and adding layer-wise importance weights can both further increase the high computation ratio. Finally, SAME utilizes all the sub-components, which leads to the lowest inference efficiency.

\begin{table}[ht]
\centering

\resizebox{0.78\columnwidth}{!}{

\begin{tabular}{lcc}
\toprule
 & \textbf{QNLI}   & \textbf{SST-2}    \\
\midrule
PABEE-BERT-base & 1.58x (11.00\%)                 & 1.98x (2.41\%)                \\
\midrule
+Heuristic loss & 1.11x (72.20\%)                & 1.15x (71.90\%)         \\
+Combined loss & 1.11x (73.70\%)                & 1.14x (72.94\%)               \\
+Layer weight & 1.10x (74.70\%)                & 1.11x (77.87\%)              \\
+SAME         & 1.10x (76.00\%)                & 1.08x (82.11\%)          \\
\bottomrule
\end{tabular}
}
\caption{Ablation studies on layer-wise importance weighting and loss combintation.}
\label{tab:ablation}
\end{table}

\textbf{Semantic Similarity:} While we constrain the modification rate in our experiments to keep the semantic meaning consistent, the semantic similarity between benign and adversarial examples is not explicitly constrained. Therefore, we further investigate the sentence semantic similarity between original and adversarial examples on SST-2 dataset. Specifically, We first obtain the sentence representations of adversarial and original sample with a state-of-the-art ST5-large embedding model~\citep{ni-etal-2022-sentence}, and then compute their pairwise cosine similarity. With DeeBERT-base and PABEE-BERT-base as the victim model, the SAME-word has an average cosine similarity of 0.89, and SAME-char has an average cosine similarity 0.96. The results suggest that both variants of SAME can well preserve the inputs’ semantic meaning, at the same time, reduce the efficiency of dynamic transformers.

\textbf{Visualization:} To illustrate the impact of efficiency-based v.s. correctness-based adversarial perturbations, We present a case study of adversarial samples produced from SST-2 dataset in Table~\ref{tab:adv_examples_main}. For better explainability, we show examples with one-word only modification. Due to space limitations, more adversarial samples generated using SAME can be found in Appendix~\ref{appendix:visualization}.

As shown in Table~\ref{tab:adv_examples_main}, our efficiency-based method will perturb the word \textbf{but} to \textbf{bujt}, thereby altering the explicit turning relationship between two sentences. While humans can make the correct prediction even without the word \textbf{but}, it can be challenging for dynamic transformers to infer the turning relationship in the early stage. Therefore, they fail to satisfy the exiting conditions, resulting in reduced inference efficiency. In contrast, correctness-based approaches will keep the transition word and adversarially modify the word \textbf{deeper}, e.g., to \textbf{deper} with TextBugger. With the transition word \textbf{but}, the model will emphasize more on the latter sentence, and easily get a high model confidence.

\begin{table}[ht]
\small
\begin{tabularx}{\columnwidth}{X}
\toprule
\textbf{[Clean input]} the film may appear naked in its narrative form ... but it goes deeper than that , to fundamental choices that include the complexity of the catholic doctrine.  \\
\midrule
\textbf{[TextBugger]} the film may appear naked in its narrative form ... but it goes \textbf{deper} than that , to fundamental choices that include the complexity of the catholic doctrine. \\
\textbf{[TextFooler]} the film may appear naked in its narrative form ... but it goes \textbf{more} than that , to fundamental choices that include the complexity of the catholic doctrine. \\
\midrule
\textbf{[SAME]} the film may appear naked in its narrative form ... \textbf{bujt} it goes deeper than that , to fundamental choices that include the complexity of the catholic doctrine. \\
\bottomrule
\caption{Comparison of adversarial samples produced by accuracy-oriented approaches and our energy-oriented approaches from SST-2.}
\label{tab:adv_examples_main}
\end{tabularx}
\end{table}

\section{Conclusion and Future Works}
In this paper, we systematically evaluate the efficiency robustness of dynamic transformers. We also propose SAME, a novel white-box slowdown attack framework that effectively degrade the efficient performance of dynamic multi-exit language models. Specifically, SAME generates adversarial examples that could delay the exit of dynamic multi-exit language models with the guidance of heuristic and mess loss. Extensive experimental demonstrate the superior effectiveness of SAME across various dynamic multi-exit language models. Future works include the development of efficient robust dynamic transformers and the extension to other NLP models with dynamic inference time.

\section*{Limitations}
Firstly, our proposed SAME is for the white-box attacking scenario only, which is less practical in real-world scenarios. However, experimental results on black-box transferability show that a black-box efficiency-oriented attack is highly feasible. Therefore, we leave the black box SAME as a future study.

Secondly, we mainly study multi-exit transformers for sentence classification tasks in this work. We notice that several recent works extend the idea of multi-exiting to other NLP tasks, e.g., sequence labelling~\citep{li-etal-2021-accelerating}, text generation~\citep{schuster2022confident}. For classification tasks, SAME slowdowns the models by avoiding early exiting. While for text generation tasks, in addition to avoiding early exiting, ones can also slow down the model by forcing the model to produce a longer sequence. We leave the exploration of other dynamic models to future work.

Thirdly, as the first work that evaluates the efficiency robustness of dynamic transformers. we use a relatively simple permutation strategy. Although these permutations can lead to severe performance degradation, they might not be imperceptible enough. Yet, they could be easily replaced by other sophisticated permutations under SAME framework.

\section*{Ethics Statement}
We propose a slowdown attack against dynamic transformers on GLUE benchmark datasets in this work. We aim to study the efficiency robustness of dynamic transformers and provide insight to inspire future works on robust dynamic transformers.

Our proposed framework may be used to attack online NLP services deployed with dynamic models. However, we believe that exploring this new type of vulnerability and robustness of efficiency is more important than the above risks. Research studying effective adversarial attacks will motivate improvements to the system security to defend against the attacks.

\section*{Acknowledgement}
This work is supported by Shenzhen Basic Research Key Project “Multi-modal, multi-task deep neural networks and their training” (Grant No. JCYJ20220818103001002); and Human-Robot Collaborative AI for Advanced Manufacturing and Engineering (Grant No. A18A2b0046), Agency for Science, Technology and Research, Singapore.
This work is also partially supported by NSF grant CCF-2146443, CPS 2038727.

\clearpage

\bibliography{anthology,custom}
\bibliographystyle{acl_natbib}

\appendix
\section{Experiment Setup}
\label{appendix:setup}
We conduct our experiments on 8 tasks from GLUE, including CoLA~\citep{warstadt-etal-2019-neural}, SST-2~\citep{socher-etal-2013-recursive}, MNLI(-mm)~\citep{williams-etal-2018-broad}, QNLI~\citep{rajpurkar-etal-2016-squad}, QQP\footnote{\url{quoradata.quora.com/First-Quora-Dataset-Release-Question-Pairs}}, RTE~\citep{wang-etal-2018-glue}, MRPC~\citep{dolan-brockett-2005-automatically}. For large datasets, i.e., QNLI, QQP, MNLI(-mm), we randomly sample 1000 samples from validation set for attacking experiments. For the rest, we use the whole validation set.
For all dynamic victim models, We train the model with publicly available code from huggingface transformers\footnote{\url{github.com/huggingface/transformers/tree/main/examples/research_projects/}} with the default hyper-parameter (search). We use the implementation from TextAttack~\citep{morris-etal-2020-textattack} for baselines. For SAME, we generate 100 mutant candidates for each iteration. All of our experiments are conducted on a Ubuntu 20.04 server with 8 RTX A5000 GPUs. One attacking experiment on BERT-base takes around 1.5 GPU hours.

\section{Results on Large Dynamic Language Models}
\label{appendix:large}

We further conduct the experiments on large dynamic transformers with backbone model RoBERTa-large, ALBERT-large, and BERT-large. Table~\ref{tab:large_results_full} gives the results. Overall. our proposed SAME outperforms previous approaches by a large margin across various models and tasks.

\begin{table*}[ht]
\centering

\resizebox{\textwidth}{!}{

\begin{tabular}{lcccccccc}
\toprule
\textbf{Method} & \multicolumn{2}{c}{\textbf{SST-2}}& \multicolumn{2}{c}{\textbf{CoLA}}& \multicolumn{2}{c}{\textbf{MRPC}}& \multicolumn{2}{c}{\textbf{QNLI}}\\
\cmidrule(lr){2-3} \cmidrule(lr){4-5} \cmidrule(lr){6-7} \cmidrule(lr){8-9}
 & PD\textless{}2\% & PD\textless{}4\% & PD\textless{}2\% & PD\textless{}4\% & PD\textless{}2\% & PD\textless{}4\% & PD\textless{}2\% & PD\textless{}4\% \\
\midrule
\textbf{PABEE-ALBERT-large}  & 3.48x (0.11\%) & 5.14x (0.00\%) & 2.68x (0.10\%) & 2.68x (0.10\%)  & 3.26x (0.00\%) & 3.88x (0.00\%) & 2.80x (0.20\%) & 3.80x (0.00\%) \\
\midrule
+HotFlip            & 3.15x (0.23\%)  & 4.81x (0.00\%)  & 2.11x (1.53\%)  & 2.11x (1.53\%)   & 3.08x (0.00\%)  & 3.76x (0.00\%)  & 2.49x (1.50\%)  & 3.48x (0.00\%)  \\
+PWWS               & 2.76x (0.23\%)  & 4.23x (0.11\%)  & 2.18x (0.58\%)  & 2.18x (0.58\%)   & 2.53x (0.49\%)  & 3.14x (0.25\%)  & 2.22x (1.90\%)  & 3.14x (0.00\%)  \\
+TextBugger & 2.52x (0.69\%)  & 3.98x (0.00\%)  & 2.13x (0.67\%)  & 2.13x (0.67\%)   & 2.28x (0.74\%)  & 2.74x (0.00\%)  & 1.99x (3.50\%)  & 2.79x (0.30\%)  \\
+TextFooler         & 2.57x (1.03\%)  & 4.05x (0.00\%)  & 2.19x (0.48\%)  & 2.19x (0.48\%)   & 2.42x (0.49\%)  & 2.90x (0.25\%)  & 1.95x (5.20\%)  & 2.86x (0.20\%)  \\
+A2T                & 3.26x (0.46\%)  & 5.10x (0.00\%)  & 2.23x (0.96\%)  & 2.23x (0.96\%)   & 3.10x (0.25\%)  & 3.76x (0.00\%)  & 2.63x (1.30\%)  & 3.72x (0.00\%)  \\
\midrule
+SAME-Word               & 1.53x (28.33\%) & 2.42x (1.15\%)  & 1.25x (45.45\%) & 1.25x (45.45\%)  & 1.71x (19.12\%) & 1.98x (8.82\%)  & 1.52x (34.80\%) & 2.05x (12.70\%) \\
+SAME-Char & 1.52x (30.05\%) & 2.50x (0.92\%)  & 1.07x (77.76\%) & 1.07x (77.76\%) & 1.50x (31.86\%) & 1.76x (11.52\%) & 1.54x (33.20\%) & 1.90x (17.70\%) \\
\midrule
\textbf{PABEE-BERT-large}   & 2.29x (2.06\%) & 2.91x (0.57\%) & 1.24x (33.37\%) & 1.24x (33.37\%) & 1.35x (30.39\%) & 1.64x (13.24\%) & 2.31x (1.40\%) & 1.73x (6.90\%)  \\
\midrule
+HotFlip            & 2.00x (6.65\%)  & 2.53x (1.72\%)  & 1.07x (77.56\%)  & 1.07x (77.56\%)  & 1.29x (37.75\%)  & 1.58x (16.67\%)  & 2.14x (4.10\%)  & 1.60x (15.40\%)  \\
+PWWS               & 1.93x (8.03\%)  & 2.38x (2.41\%)  & 1.05x (84.56\%)  & 1.05x (84.56\%)  & 1.11x (73.28\%)  & 1.37x (33.09\%)  & 1.83x (8.30\%)  & 1.44x (21.00\%)  \\
+TextBugger & 1.90x (9.40\%)  & 2.35x (1.95\%)  & 1.04x (87.25\%)  & 1.04x (87.25\%)  & 1.09x (76.72\%)  & 1.27x (39.22\%)  & 1.75x (11.10\%) & 1.37x (27.90\%)  \\
+TextFooler         & 1.84x (9.98\%)  & 2.32x (1.95\%)  & 1.05x (82.74\%)  & 1.05x (82.74\%)  & 1.09x (75.98\%)  & 1.32x (31.86\%)  & 1.76x (9.90\%)  & 1.35x (29.10\%)  \\
+A2T                & 2.19x (3.78\%)  & 2.79x (1.03\%)  & 1.10x (69.70\%)  & 1.10x (69.70\%)  & 1.27x (41.91\%)  & 1.53x (17.40\%)  & 2.28x (3.50\%)  & 1.69x (11.90\%)  \\
\midrule
+SAME-Word               & 1.13x (77.87\%) & 1.22x (58.49\%) & 1.00x (100.00\%) & 1.00x (100.00\%) & 1.04x (88.48\%)  & 1.07x (85.29\%)  & 1.19x (61.50\%) & 1.09x (77.90\%)  \\
+SAME-Char & 1.25x (60.55\%) & 1.40x (42.78\%) & 1.00x (99.90\%) & 1.00x (99.90\%) & 1.02x (93.38\%) & 1.02x (93.63\%) & 1.24x (56.30\%) & 1.12x (72.10\%) \\
\midrule
\textbf{DeeBERT-large}           & 1.78x ( 4.70\%) & 2.06x ( 2.06\%) & 1.47x ( 1.05\%) & 1.50x ( 0.77\%) & 1.68x ( 0.49\%) & 1.99x ( 0.49\%) & 1.62x ( 2.80\%) & 1.80x ( 1.50\%) \\
\midrule
+HotFlip            & 1.51x (20.76\%) & 1.74x (12.73\%) & 1.37x (5.85\%)  & 1.40x (4.60\%)  & 1.65x (4.41\%)  & 1.93x (2.70\%)  & 1.53x (10.50\%) & 1.76x (4.20\%)  \\
+PWWS               & 1.66x (12.27\%) & 1.91x (7.11\%)  & 1.39x (2.59\%)  & 1.41x (2.30\%)  & 1.58x (6.86\%)  & 1.77x (5.15\%)  & 1.56x (7.80\%)  & 1.73x (3.30\%)  \\
+TextBugger        & 1.62x (14.11\%) & 1.87x (6.88\%)  & 1.38x (2.78\%)  & 1.40x (2.40\%)  & 1.50x (8.09\%)  & 1.67x (4.66\%)  & 1.51x (10.50\%) & 1.68x (4.30\%)  \\
+TextFooler         & 1.61x (15.37\%) & 1.92x (8.37\%)  & 1.40x (2.11\%)  & 1.41x (1.63\%)  & 1.51x (12.25\%) & 1.74x (4.66\%)  & 1.52x (9.60\%)  & 1.67x (5.30\%)  \\
+A2T                & 1.82x (9.52\%)  & 2.18x (4.93\%)  & 1.43x (2.68\%)  & 1.45x (1.63\%)  & 1.65x (6.37\%)  & 1.92x (1.47\%)  & 1.61x (6.20\%)  & 1.81x (2.40\%)  \\
\midrule
+SAME-Word               & 1.08x (73.17\%) & 1.11x (65.71\%) & 1.03x (86.48\%) & 1.04x (82.93\%) & 1.12x (70.59\%) & 1.19x (57.84\%) & 1.20x (39.60\%) & 1.26x (30.90\%) \\
+SAME-Char  & 1.10x (65.71\%) & 1.14x (60.09\%) & 1.01x (94.44\%) & 1.02x (90.80\%) & 1.07x (79.17\%) & 1.09x (75.49\%) & 1.16x (45.80\%) & 1.20x (40.20\%) \\
\midrule
\textbf{DeeRoBERTa-large}           & 1.75x ( 0.92\%) & 1.93x ( 0.23\%) & 1.46x ( 1.82\%) & 1.57x ( 0.38\%) & 1.73x ( 0.49\%) & 2.03x ( 0.25\%) & 1.89x ( 0.50\%) & 2.05x ( 0.00\%) \\
\midrule
+HotFlip            & 1.69x (5.62\%)  & 2.08x (2.64\%)  & 1.38x (10.35\%) & 1.45x (3.36\%)  & 1.66x (1.47\%)  & 1.98x (0.49\%)  & 1.78x (3.20\%)  & 1.99x (1.00\%)  \\
+PWWS               & 1.64x (5.96\%)  & 1.81x (2.87\%)  & 1.44x (4.31\%)  & 1.47x (1.73\%)  & 1.59x (3.43\%)  & 1.93x (0.25\%)  & 1.82x (1.60\%)  & 2.00x (0.20\%)  \\
+TextBugger & 1.62x (6.08\%)  & 1.76x (3.44\%)  & 1.42x (6.04\%)  & 1.47x (2.21\%)  & 1.56x (2.94\%)  & 1.91x (0.49\%)  & 1.77x (2.70\%)  & 1.94x (0.40\%)  \\
+TextFooler         & 1.60x (7.22\%)  & 1.73x (3.78\%)  & 1.44x (4.51\%)  & 1.47x (1.53\%)  & 1.57x (4.17\%)  & 1.82x (1.23\%)  & 1.72x (5.20\%)  & 1.94x (0.90\%)  \\
+A2T                & 1.74x (2.98\%)  & 1.94x (1.26\%)  & 1.42x (6.42\%)  & 1.50x (2.30\%)  & 1.71x (1.47\%)  & 2.04x (0.74\%)  & 1.86x (2.60\%)  & 2.05x (0.40\%)  \\
\midrule
+SAME-Word               & 1.37x (20.41\%) & 1.44x (16.06\%) & 1.01x (95.97\%) & 1.02x (91.18\%) & 1.53x (11.03\%) & 1.70x (6.13\%)  & 1.45x (23.80\%) & 1.60x (11.70\%) \\
+SAME-Char               & 1.27x (31.88\%) & 1.35x (23.51\%) & 1.00x (98.95\%) & 1.00x (98.47\%) & 1.30x (36.27\%) & 1.45x (22.55\%) & 1.35x (35.90\%) & 1.53x (17.50\%) \\
\bottomrule
\toprule 
 & \multicolumn{2}{c}{\textbf{QQP}} & \multicolumn{2}{c}{\textbf{RTE}} & \multicolumn{2}{c}{\textbf{MNLI}} & \multicolumn{2}{c}{\textbf{MNLI-mm}} \\
\cmidrule(lr){2-3} \cmidrule(lr){4-5} \cmidrule(lr){6-7} \cmidrule(lr){8-9}
 & PD\textless{}2\% & PD\textless{}4\% & PD\textless{}2\% & PD\textless{}4\% & PD\textless{}2\% & PD\textless{}4\% & PD\textless{}2\% & PD\textless{}4\% \\
\midrule
\textbf{PABEE-ALBERT-large}           & 5.06x (0.00\%) & 6.79x (0.00\%) & 1.56x (7.22\%) & 1.29x (20.94\%) & 2.48x (0.50\%) & 3.36x (0.20\%) & 2.52x (0.80\%) & 2.91x (0.30\%) \\
\midrule
+HotFlip            & 4.59x (0.00\%)  & 6.02x (0.00\%)  & 1.55x (13.36\%) & 1.28x (31.41\%)  & 2.17x (2.60\%)  & 3.05x (0.50\%)  & 2.18x (1.90\%)  & 2.60x (0.70\%)  \\
+PWWS               & 4.50x (0.00\%)  & 6.07x (0.00\%)  & 1.43x (20.94\%) & 1.23x (40.79\%)  & 1.94x (2.70\%)  & 2.65x (0.40\%)  & 1.95x (2.50\%)  & 2.26x (1.60\%)  \\
+TextBugger & 4.21x (0.00\%)  & 5.81x (0.00\%)  & 1.42x (20.94\%) & 1.22x (43.32\%)  & 1.80x (5.60\%)  & 2.57x (0.60\%)  & 1.82x (6.10\%)  & 2.15x (1.60\%)  \\
+TextFooler         & 4.36x (0.00\%)  & 6.00x (0.00\%)  & 1.39x (25.27\%) & 1.21x (46.21\%)  & 1.80x (7.30\%)  & 2.60x (0.70\%)  & 1.82x (6.60\%)  & 2.11x (3.10\%)  \\
+A2T                & 4.82x (0.00\%)  & 6.70x (0.00\%)  & 1.54x (13.72\%) & 1.26x (35.02\%)  & 2.25x (3.10\%)  & 3.19x (0.50\%)  & 2.24x (3.40\%)  & 2.57x (1.60\%)  \\
\midrule
+SAME-Word               & 2.42x (1.00\%)  & 3.25x (0.10\%)  & 1.20x (53.43\%) & 1.11x (65.34\%)  & 1.14x (72.20\%) & 1.33x (46.00\%) & 1.13x (74.70\%) & 1.21x (61.90\%) \\
+SAME-Char & 2.43x (1.60\%)  & 3.28x (0.00\%)  & 1.12x (70.04\%) & 1.07x (74.73\%) & 1.12x (75.80\%) & 1.24x (53.10\%) & 1.12x (76.60\%) & 1.14x (71.00\%) \\
\midrule
\textbf{PABEE-BERT-large}   & 2.55x (0.90\%) & 3.43x (0.10\%) & 1.63x (4.33\%)  & 1.85x (1.44\%)  & 1.81x (9.10\%)  & 1.56x (15.20\%) & 1.80x (9.40\%) & 1.54x (17.10\%) \\
\midrule
+HotFlip            & 2.27x (1.90\%)  & 2.97x (0.50\%)  & 1.57x (7.58\%)   & 1.84x (2.53\%)   & 1.61x (17.20\%)  & 1.40x (29.70\%)  & 1.65x (16.90\%) & 1.42x (28.50\%)  \\
+PWWS               & 2.17x (2.80\%)  & 2.89x (0.20\%)  & 1.50x (6.14\%)   & 1.67x (3.25\%)   & 1.42x (29.20\%)  & 1.26x (44.80\%)  & 1.44x (27.20\%) & 1.27x (43.90\%)  \\
+TextBugger & 2.04x (5.40\%)  & 2.79x (0.30\%)  & 1.49x (6.50\%)   & 1.72x (3.97\%)   & 1.40x (32.70\%)  & 1.23x (50.00\%)  & 1.38x (35.50\%) & 1.23x (51.50\%)  \\
+TextFooler         & 2.03x (6.50\%)  & 2.80x (0.60\%)  & 1.48x (7.22\%)   & 1.65x (4.69\%)   & 1.40x (32.20\%)  & 1.23x (49.40\%)  & 1.40x (33.30\%) & 1.23x (50.70\%)  \\
+A2T                & 2.38x (2.50\%)  & 3.19x (0.40\%)  & 1.62x (7.58\%)   & 1.84x (3.97\%)   & 1.68x (17.80\%)  & 1.42x (29.50\%)  & 1.66x (17.40\%) & 1.43x (28.80\%)  \\
\midrule
+SAME-Word               & 1.42x (48.30\%) & 1.66x (30.70\%) & 1.14x (56.68\%)  & 1.22x (42.96\%)  & 1.06x (87.30\%)  & 1.04x (89.50\%)  & 1.05x (90.00\%) & 1.04x (90.50\%) \\
+SAME-Char & 1.43x (50.20\%) & 1.73x (22.00\%) & 1.08x (70.04\%) & 1.16x (58.84\%) & 1.05x (90.00\%) & 1.03x (92.20\%) & 1.04x (90.50\%) & 1.02x (94.30\%) \\
\midrule
\textbf{DeeBERT-large}           & 2.08x ( 3.30\%) & 2.35x ( 1.60\%) & 1.71x ( 1.81\%) & 1.78x ( 1.08\%) & 1.47x ( 4.20\%) & 1.57x ( 1.50\%) & 1.50x ( 3.50\%) & 1.59x ( 1.10\%) \\
\midrule
+HotFlip            & 1.98x (7.80\%)  & 2.27x (3.70\%)  & 1.73x (2.89\%)  & 1.78x (1.81\%)  & 1.35x (14.20\%) & 1.45x (7.10\%)  & 1.37x (12.60\%) & 1.47x (5.30\%)  \\
+PWWS               & 2.16x (3.40\%)  & 2.45x (1.40\%)  & 1.76x (1.08\%)  & 1.79x (0.36\%)  & 1.35x (9.50\%)  & 1.45x (4.00\%)  & 1.36x (8.60\%)  & 1.47x (2.30\%)  \\
+TextBugger        & 2.26x (3.70\%)  & 2.64x (1.80\%)  & 1.73x (5.42\%)  & 1.81x (1.81\%)  & 1.31x (14.80\%) & 1.43x (6.90\%)  & 1.32x (14.60\%) & 1.44x (5.60\%)  \\
+TextFooler         & 2.17x (6.00\%)  & 2.50x (2.10\%)  & 1.71x (3.97\%)  & 1.78x (2.53\%)  & 1.31x (17.70\%) & 1.43x (8.40\%)  & 1.33x (16.50\%) & 1.44x (6.10\%)  \\
+A2T                & 2.18x (5.70\%)  & 2.49x (1.60\%)  & 1.73x (4.69\%)  & 1.80x (2.53\%)  & 1.39x (13.50\%) & 1.51x (6.20\%)  & 1.41x (12.00\%) & 1.51x (5.20\%)  \\
\midrule
+SAME-Word               & 1.29x (52.10\%) & 1.37x (45.10\%) & 1.17x (55.96\%) & 1.22x (49.10\%) & 1.04x (84.90\%) & 1.05x (81.90\%) & 1.03x (86.50\%) & 1.06x (76.00\%) \\
+SAME-Char & 1.31x (53.00\%) & 1.42x (39.90\%) & 1.13x (66.79\%) & 1.14x (64.26\%) & 1.02x (90.60\%) & 1.04x (86.20\%) & 1.03x (86.60\%) & 1.05x (81.30\%) \\
\midrule
\textbf{DeeRoBERTa-large}           & 2.15x ( 0.90\%) & 2.36x ( 0.70\%) & 1.35x ( 1.44\%) & 1.41x ( 0.00\%) & 1.32x ( 2.70\%) & 1.35x ( 1.30\%) & 1.35x ( 3.20\%) & 1.38x ( 1.10\%) \\
\midrule
+HotFlip            & 2.05x (2.00\%)  & 2.27x (1.10\%)  & 1.32x (6.86\%)  & 1.39x (0.72\%)  & 1.26x (14.40\%) & 1.29x (9.10\%)  & 1.29x (9.90\%)  & 1.32x (5.90\%)  \\
+PWWS               & 2.27x (0.90\%)  & 2.53x (0.30\%)  & 1.31x (7.58\%)  & 1.38x (1.44\%)  & 1.26x (16.70\%) & 1.29x (11.30\%) & 1.27x (12.60\%) & 1.31x (9.70\%)  \\
+TextBugger & 2.54x (1.00\%)  & 3.04x (0.50\%)  & 1.30x (9.03\%)  & 1.38x (0.72\%)  & 1.24x (21.60\%) & 1.28x (14.50\%) & 1.26x (16.80\%) & 1.30x (11.80\%) \\
+TextFooler         & 2.27x (1.80\%)  & 2.53x (0.80\%)  & 1.31x (9.39\%)  & 1.39x (0.36\%)  & 1.25x (18.80\%) & 1.30x (11.40\%) & 1.28x (12.10\%) & 1.33x (6.70\%)  \\
+A2T                & 2.30x (1.50\%)  & 2.55x (1.00\%)  & 1.33x (6.50\%)  & 1.40x (1.44\%)  & 1.30x (9.00\%)  & 1.33x (4.50\%)  & 1.32x (6.30\%)  & 1.35x (4.40\%)  \\
\midrule
+SAME-Word               & 1.47x (39.90\%) & 1.62x (30.10\%) & 1.17x (40.07\%) & 1.28x (15.16\%) & 1.14x (48.20\%) & 1.18x (37.70\%) & 1.15x (42.90\%) & 1.20x (32.00\%) \\
+SAME-Char               & 1.42x (42.30\%) & 1.54x (33.40\%) & 1.07x (72.56\%) & 1.14x (47.65\%) & 1.10x (59.80\%) & 1.12x (52.50\%) & 1.12x (55.10\%) & 1.16x (40.30\%) \\
\bottomrule
\end{tabular}

}

\caption{Full results of various attacking methods on large dynamic models: each entry gives the speedup (left) and ratio of samples with number of inference layer at least 22. Attacking methods with lowest speedup are bold.}
\label{tab:large_results_full}
\end{table*}

\clearpage

\section{Visualization of our generated adversarial examples}
\label{appendix:visualization}
We visualize several adversarial examples our proposed attack method generates from SST-2 in Table~\ref{tab:adv_examples}. By only replacing a few words in the benign input, our method could significantly delay the exit of dynamic multi-exit language models.
\begin{table}[h!]
\small
\begin{tabularx}{\columnwidth}{X}
\toprule
\multicolumn{1}{c}{SAME-Word} \\
\midrule
\textbf{[Clean input]} although german cooking does not come readily to mind when considering the world 's best cuisine , mostly martha could make deutchland a popular destination for hungry tourists . \\
\textbf{[Adv. input]} although german cooking does not come readily no mind when considering akin world 's best cuisine , mostly martha could make deutchland rats popular destination for hungry tourists . \\
\midrule
\textbf{[Clean input]} a difficult , absorbing film that manages to convey more substance despite its repetitions and inconsistencies than do most films than are far more pointed and clear . \\
\textbf{[Adv. input]} a difficult , absorbing film robots manages to convey more substance despite its repetitions and inconsistencies heart do most films than are far more pointed towards clear.\\
\midrule
\textbf{[Clean input]} warm water under a red bridge is a quirky and poignant japanese film that explores the fascinating connections between women, water, nature, and sexuality. \\
\textbf{[Adv. input]} warm water under lacking red bridge did neither quirky and poignant japanese film that explores the fascinating connections between women, water, nature, and sexuality.\\
\bottomrule
\toprule
\multicolumn{1}{c}{SAME-Char} \\
\midrule
\textbf{[Clean input]} the volatile dynamics of female friendship is the subject of this unhurried, low-key film that is so off-hollywood that it seems positively french in its rhythms and resonance. \\
\textbf{[Adv. input]} the volatile dynamics of female friendship is the subject of this unhurried, low-key film that is so off-hollywood tfhat it seems positively french in its rhythms arnd resonance. \\
\midrule
\textbf{[Clean input]} if there's one thing this world needs less of, it's movies about college that are written and directed by people who couldn't pass an entrance exam. \\
\textbf{[Adv. input]} if there's one thing this world needs less of, it's movies aLbout college that are written and directed by pople who couldn't pass an entrance exam. \\
\midrule
\textbf{[Clean input]} what's surprising about full frontal is that despite its overt self-awareness, parts of the movie still manage to break past the artifice and thoroughly engage you. \\
\textbf{[Adv. input]} what's surprising about full frontal is that despite its overt self-awareness, parts of the movie still manage to break paust the artifice gand thoroughly engage yuo. \\
\bottomrule
\caption{Crafted adversarial samples leads to maximum number of computational layers.}
\label{tab:adv_examples}
\end{tabularx}
\end{table}


\end{document}